\documentclass[conference]{IEEEtran}
\IEEEoverridecommandlockouts

\usepackage{amsmath,amssymb,amsfonts}
\usepackage{graphicx}
\usepackage{textcomp}
\usepackage{xcolor}
\usepackage{bbm}
\usepackage{enumitem}
\usepackage{tabularx, booktabs}
\usepackage[style=ieee, maxcitenames=2, mincitenames=1, natbib=true, dashed=false, maxbibnames=99, minbibnames=99]{biblatex}
\usepackage{soul,color}
\usepackage{siunitx}
\usepackage{hyperref}
\usepackage[capitalise]{cleveref}
\usepackage{dsfont}
\usepackage{minted}
\usepackage[caption=false]{subfig}
\usepackage[linesnumbered, ruled, vlined]{algorithm2e}
\usepackage{float}
\usepackage{pifont}
\SetAlFnt{\small}
\newcommand{\xmark}{\ding{55}}
\newcommand{\cmark}{\ding{51}}
\crefname{algocf}{alg.}{algs.}
\Crefname{algocf}{Algorithm}{Algorithms}
\DeclareMathOperator*{\argmin}{argmin}

\usepackage{setspace}
\usepackage{multirow}

\begin{document}

\title{
    \LARGE \bf Priority-Aware Multi-Robot Coverage Path Planning
    
    \author{Kanghoon Lee, Hyeonjun Kim, Jiachen Li, Jinkyoo Park}
    
    \thanks{K. Lee is with the Korea Advanced Institute of Science and Technology (KAIST), Daejeon, South Korea. {\tt\small leehoon@kaist.ac.kr}.}
    \thanks{H. Kim is with the Korea Military Academy (KMA), Seoul, South Korea. {\tt\small hyunjoon0605@kma.ac.kr}.}
    \thanks{J. Li is with the University of California, Riverside (UCR), CA, USA. {\tt\small jiachen.li@ucr.edu}.}
    \thanks{J. Park is with the Korea Advanced Institute of Science and Technology (KAIST) and Omelet, South Korea. {\tt\small jinkyoo.park@kaist.ac.kr}.}
}
\maketitle

\begin{abstract}

Multi-robot systems are widely used for coverage tasks that require efficient coordination across large environments. In Multi-Robot Coverage Path Planning (MCPP), the objective is typically to minimize the makespan by generating non-overlapping paths for full-area coverage. However, most existing methods assume uniform importance across regions, limiting their effectiveness in scenarios where some zones require faster attention. We introduce the Priority-Aware MCPP (PA-MCPP) problem, where a subset of the environment is designated as prioritized zones with associated weights. The goal is to minimize, in lexicographic order, the total priority-weighted latency of zone coverage and the overall makespan. To address this, we propose a scalable two-phase framework combining (1) greedy zone assignment with local search, spanning-tree-based path planning, and (2) Steiner-tree-guided residual coverage. Experiments across diverse scenarios demonstrate that our method significantly reduces priority-weighted latency compared to standard MCPP baselines, while maintaining competitive makespan. Sensitivity analyses further show that the method scales well with the number of robots and that zone coverage behavior can be effectively controlled by adjusting priority weights.

\end{abstract}

\section{Introduction}
    \label{sec:intro}

Multi-robot systems are increasingly used to tackle complex spatial and temporal tasks that exceed the capabilities of a single robot \cites{zhang2025lamma, kim2025human, gaostamp, li2025comamba}. Operating in parallel enables faster task completion and scalability to large and complex environments. 
Such systems are applied in diverse domains, including search and rescue \cite{ai2021coverage, wu2024autonomous}, surveillance \cite{velhal2022decentralized, bajaj2024multivehicle}, monitoring \cite{asghar2024multi}, warehouse logistics \cite{bolu2021adaptive}, motion/occupancy prediction \cite{wang2025cmp,wang2025deployable,wang2025uniocc}, and object transport \cite{wang2018cooperative}. However, realizing the full potential of multi-robot systems requires solving challenging coordination problems, particularly when task objectives involve spatial coverage \cite{tang2025large}. 

A prominent class of such structured tasks involves coverage, where robots are required to collectively visit or observe every region of a known environment \cite{hazon2005redundancy}. In MCPP, the main objective is to minimize the overall makespan by generating coordinated, non-overlapping paths that enable the robots to cover the entire region efficiently \cite{dong2020artificially, tang2021mstc, tang2023mixed}. In some settings, reducing the number of turns are also considered to improve motion smoothness or energy efficiency \cite{vandermeulen2019turn, lu2023tmstc}. 
While MCPP has been widely explored, most approaches assume uniform importance across all regions. 
This assumption becomes inadequate when certain zones demand faster attention, such as hazardous areas, critical assets, or time-sensitive inspection sites. Delayed coverage in these regions can degrade task performance or safety, underscoring the need to incorporate spatial priorities in path planning.

\begin{figure}[t!]
\vspace{-7pt}
\centering
\subfloat[MCPP$\quad$]{
  \includegraphics[width=0.22\textwidth]{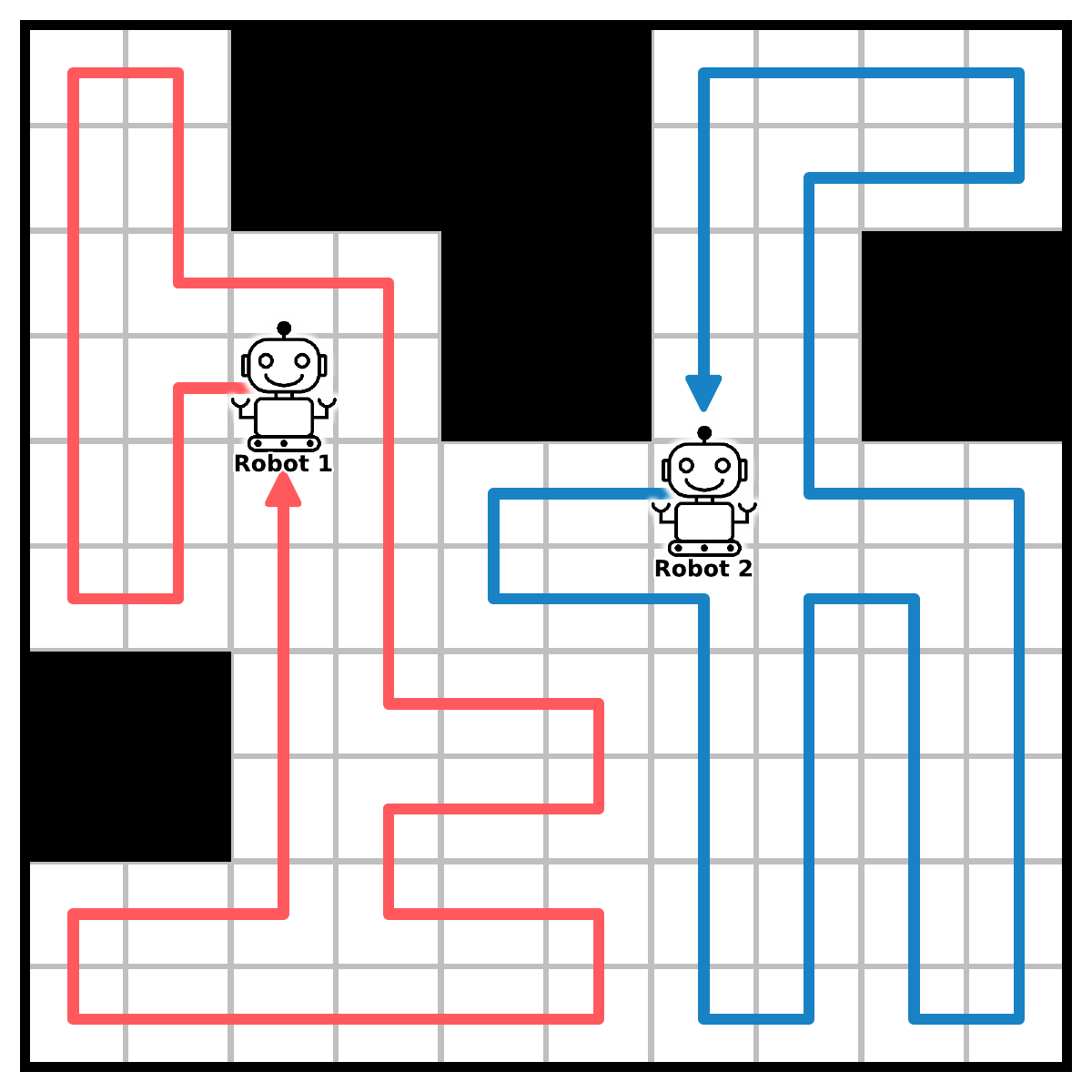}
  \label{fig:teaser1}
}
\subfloat[Priority-Aware MCPP$\quad$]{
  \includegraphics[width=0.22\textwidth]{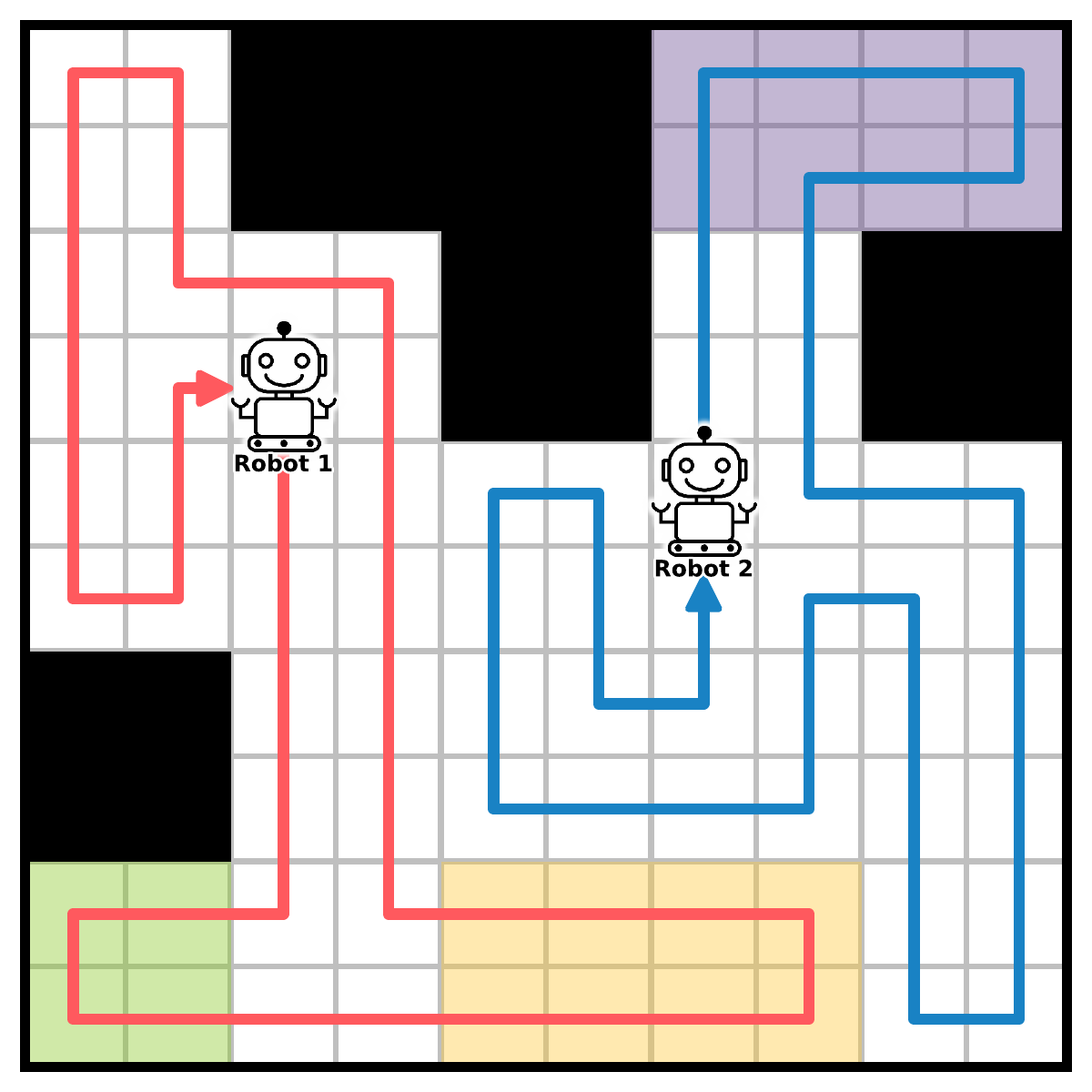}
  \label{fig:teaser2}
}
\caption{\textbf{Comparison of MCPP and Priority-Aware MCPP.} While standard MCPP minimizes overall makespan by distributing coverage among robots, Priority-Aware MCPP additionally accounts for spatial priority by encouraging early completion of high-priority zones. Prioritized zones are indicated by green, yellow, and purple shaded areas in the right figure.}
\label{fig:teaser}
\vspace{10pt}
\end{figure}

To address the limitations of uniform coverage, we define the Priority-Aware MCPP (PA-MCPP) problem, where certain regions are assigned priority weights reflecting their urgency or value. The objective is to plan robot paths that cover the entire environment while ensuring that high-priority zones are visited as early as possible, reflecting safety-critical missions where strict priority compliance is essential \cite{kusnur2021planning, song2022multi}.
Specifically, we minimize two criteria in lexicographic order: (1) the priority-weighted latency, defined as the time it takes to fully cover each zone multiplied by its priority weight, and (2) the makespan, defined as the time when the last robot completes its coverage task as shown in \Cref{fig:teaser}. 
This lexicographic objective makes a direct application of existing MCPP methods challenging, as they lack the mechanism to first minimize latency and then re-balance the remaining coverage. It is motivated by the practical need to prioritize critical zones, even at the cost of a slightly longer makespan. Unlike weighted-sum approaches, the lexicographic objective enforces this strict priority structure, ensuring covered first.

To solve the PA-MCPP problem, we propose a two-phase framework aligned with the lexicographic objective. In the first phase, prioritized zones are assigned to robots using a greedy heuristic minimizing priority-weighted latency, followed by local search refinement. Each robot then generates its trajectory by sequentially covering assigned zones via zone-wise spanning tree construction and traversal. In the second phase, full coverage is ensured by building a Steiner tree over the remaining areas and distributing it among robots using a workload-balancing scheme based on prior effort. The final paths integrate results from both phases, achieving both early coverage of critical zones and efficient overall coverage.

In summary, our main contributions are as follows:
\begin{itemize}
    \item We formalize the PA-MCPP problem with lexicographic objectives that prioritize early coverage of critical regions while ensuring complete environment coverage.
    \item We develop a scalable two-phase framework that integrates greedy zone assignment with local search refinement for prioritized zones, and Steiner-tree-guided residual coverage with workload balancing.
    \item We validate our method on diverse layouts, achieving significant gains in priority-weighted latency while maintaining competitive makespan against MCPP baselines.
\end{itemize}

\section{Related Works}
    \label{sec:related_works}

\subsection{Multi-Robot Coverage Path Planning (MCPP)}
MCPP aims to generate coordinated paths for multiple robots to fully cover an environment while minimizing makespan and avoiding redundancy. Divide-and-conquer strategies, from early decentralized partitioning methods \cite{jager2002dynamic} to more recent approaches such as DARP \cite{kapoutsis2017darp} and GM-VPC \cite{nair2020gm}, partition the environment into balanced, non-overlapping regions before computing coverage paths. Spanning tree coverage (STC) \cite{gabriely2001spanning}, a single-robot method, generates continuous coverage paths based on a spanning tree. Its multi-robot extension, including MSTC \cite{hazon2005redundancy}, MFC \cite{zheng2005multi}, AWSTC \cite{dong2020artificially}, MSTC$^*$ \cite{tang2021mstc}, and MIP-SRH \cite{tang2023mixed} construct spanning trees or forests to generate coverage paths with reduced overlap. To handle large-scale numbers of robots and practical conflict resolution, LS-MCPP \cite{tang2025large} incorporates local search and path deconfliction to avoid robot-robot collisions. 
Turn-minimization methods such as TMC~\cite{vandermeulen2019turn} and TMSTC$^*$~\cite{lu2023tmstc} reduce turns for smoother motion, and a refinement based on local search for a single robot is explored \cite{krupke2024near}.
For partially known or dynamic environments, online methods like ConCPP \cite{mitra2024online} generate paths incrementally during execution. 
While prior works focus on balanced workload, collision avoidance, and motion efficiency, they assume uniform importance across all regions. In contrast, our PA-MCPP performs local search at a higher level for zone assignment, integrating lexicographic objectives to prioritize early coverage of high-weight zones. 

\subsection{Task Planning with Regional Priorities}
Task planning with regional priorities addresses scenarios where some areas require earlier attention than others, a situation not handled by uniform coverage approaches. Classical formulations such as the Traveling Repairman Problem (TRP, \cite{afrati1986complexity}) and its weighted and multi-depot variants explicitly minimize the weighted sum of arrival times to targets, ensuring high-importance locations are serviced promptly \cite{blum1994minimum, bruni2022multi}. Beyond these spatial formulations, persistent monitoring studies also model time-varying priorities through information decay \cite{lan2013planning, han2021age}.
Extensions of these ideas appear in coverage and inspection tasks, where priority maps or region weights guide agents to valuable areas earlier in the mission \cite{kusnur2021planning, song2022multi, poudel2023priority}. Multi-robot informative path planning similarly emphasizes revisiting outdated areas to preserve sensing quality over time \cite{Singh2007efficient}. Similarly, in maritime Search and Rescue (SAR), the task often involves locating targets in high-probability regions based on drift predictions and environmental constraints, aiming to reduce path overlap and travel time \cite{ai2021coverage, wu2024autonomous}. While these works improve responsiveness to prioritized regions, they do not ensure complete coverage. Our approach enforces both full coverage and prioritization through a lexicographic objective.

\section{Problem Formulation}
    \label{sec:problem_formulation}

We consider a PA-MCPP problem defined over a two-dimensional grid environment. The environment is modeled as an undirected graph $G = (V, E)$, where each node $v \in V$ represents a discrete cell in the grid. 
Each vertex $v$ is assigned a traversal cost $c(v)$.
An edge $e \in E$ exists between two nodes if they are horizontally or vertically adjacent. To facilitate STC-based path generation, we introduce a hypergraph representation $H = (V_h, E_h)$ derived from $G$. Specifically, each $2 \times 2$ block of grid cells in $V$ is contracted into a single hypervertex in $V_h$, and a hyperedge in $E_h$ connects two hypervertices if at least one pair of their constituent grid cells is adjacent in $G$. 
The environment contains a set of $n$ zones, $Z = \{ Z_j \}_{j=1}^n$, where each zone $Z_j \subseteq V$ is associated with a positive priority weight $w_j > 0$ for all $j \in \{1, \dots, n\}$. All nodes within a zone are assumed to be connected. Let $R = \{ r_i \}_{i=1}^k \subseteq V$ denote the initial positions of $k$ robots, indexed by the robot set $I = \{1, 2, \dots, k\}$. Then, a PA-MCPP instance is represented by the tuple $(G, Z, R, I)$.

For each robot $i \in I$, a path is defined as a sequence of nodes $\pi_i = (v_1, v_2, \dots, v_{|\pi_i|})$ such that the path starts and ends at the initial position of robot, i.e., $v_1 = v_{|\pi_i|} = r_i$, and each consecutive node pair $(v_{j-1}, v_j)$ is an edge in $E$ for all $j = 2, 3, \dots, |\pi_i|$. Let $\pi_i^{(t)} = \{ v_1, v_2, \dots, v_t \}$ denote the set of nodes visited by robot $i$ up to timestep $t \leq |\pi_i|$. The coverage time of zone $Z_j$, denoted $T_j$, is the earliest timestep when all its nodes are visited by at least one robot:
\begin{equation}
    T_j = \min \left\{ t \mid Z_j \subseteq \bigcup_{i \in I} \pi_i^{(t)} \right\}, \quad \forall j \in \{1, \dots, n\}.
\end{equation}

The goal of PA-MCPP is to find a set of robot paths that optimizes a lexicographic objective, where multiple criteria are prioritized and one solution dominates another if it performs better on a higher-priority objective:
\begin{align}
    \underset{\{\pi_i\}_{i\in{I}}}{\text{lex min}} \quad & \sum_{j=1}^{n} w_j T_j,\; \max_{i \in I} |\pi_i| \label{eq:2} \\
    \text{s.t.} \quad & \bigcup_{i \in I} \pi_i = V. \label{eq:3}
\end{align}
The first objective is to minimize the total priority-weighted latency of zone coverage. The second objective minimizes the makespan, defined as the maximum path length among all robots. While a weighted-sum formulation can express similar trade-offs, its weights are difficult to set consistently across different environments.
If no prioritized zones exist, the PA-MCPP problem reduces to the standard MCPP~\cite{tang2025large}, whose notation and structure we adopt for better consistency.

\section{Methods}
    \label{sec:methods}

\begin{figure*}[tb!]
    \centering
    \includegraphics[width=\textwidth]{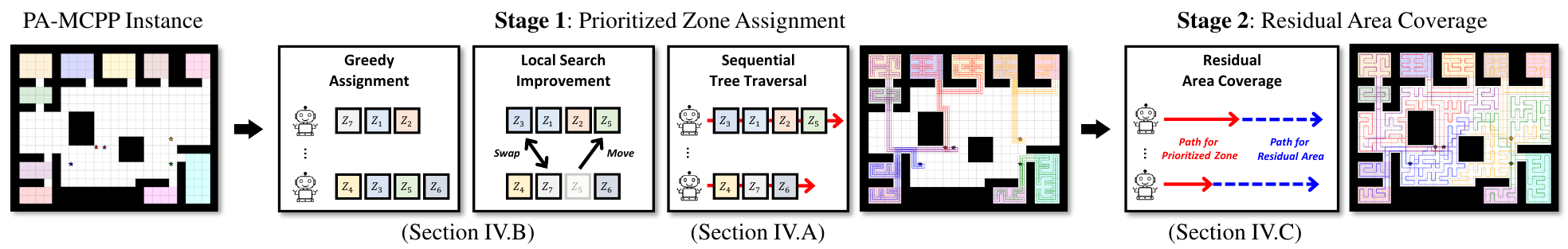}
    \vspace{-20pt}
    \caption{\textbf{Overview of the PA-MCPP algorithm.} The instance consists of maps with prioritized zones and associated weights. Stage 1 assigns prioritized zones to robots using a greedy allocation followed by local search to refine assignments, then plans traversal sequences within each zone. Stage 2 covers remaining areas to ensure complete coverage, balancing workloads based on previous assignments to minimize makespan.}
    \vspace{-7pt}
    \label{fig:method_overview}
\end{figure*}

We propose a two-phase method to solve the PA-MCPP problem, explicitly grounded in the STC paradigm. 
The overall approach aligns with the lexicographic objective: the first phase minimizes the total priority-weighted latency of zone coverage, and the second ensures full coverage of the environment, as shown in \Cref{fig:method_overview}. 
Constructing multiple coverage trees is related to the min–max tree cover problem \cite{even2004min}, but we build them to prioritize zones and then generate trees for full coverage.
We first describe the single-robot path generation for multiple assigned zones in a given order in \Cref{subsec:method_1}, then extend it to the multi-robot case with greedy and local search zone assignment in \Cref{subsec:method_2}. Finally, residual areas are covered by constructing and concatenating residual coverage trees with prioritized trajectories in \Cref{subsec:method_3}.
For clarity of presentation, we use the symbol $u$ to denote a vertex in the underlying graph used for path planning, which in our case is a hypervertex $u \in V_h$ of the hypergraph $H = (V_h, E_h)$. The cost of a hypervertex is the average cost of its four constituent grid cells, and the cost of a hyperedge is the average of the costs of the two hypervertices it connects \cite{tang2023mixed}.

\subsection{Single Robot Path Planning with Multiple Zones}\label{subsec:method_1}

\begin{figure}[t!]
\centering
\subfloat[Zone-Wise Tree Construction$\quad$]{
  \includegraphics[width=0.22\textwidth]{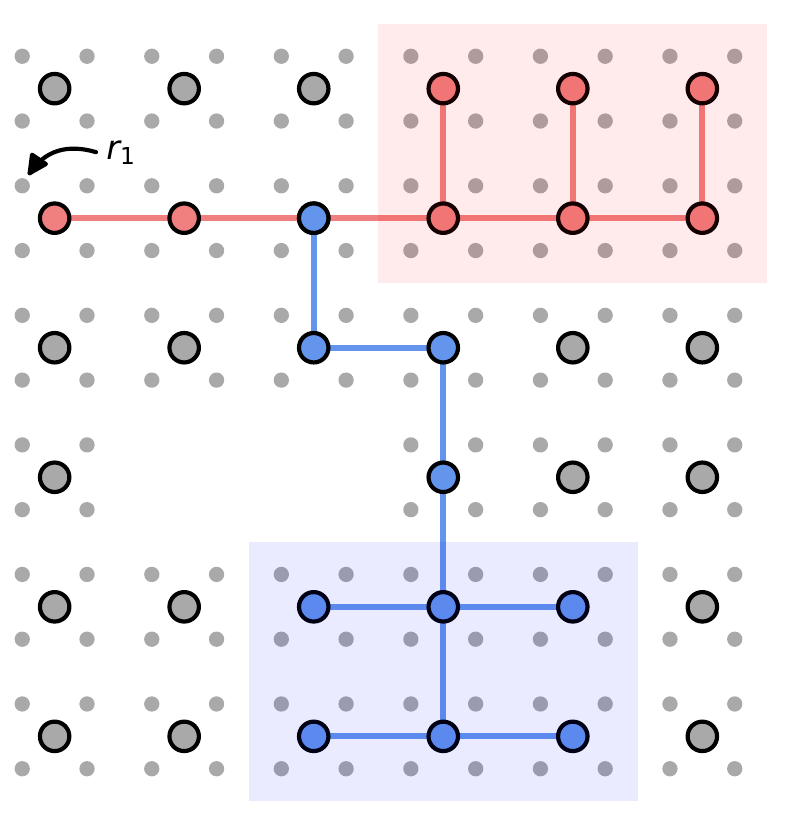}
  \label{fig:method_a_a}
}
\subfloat[Sequential Tree Traversal$\quad$]{
  \includegraphics[width=0.22\textwidth]{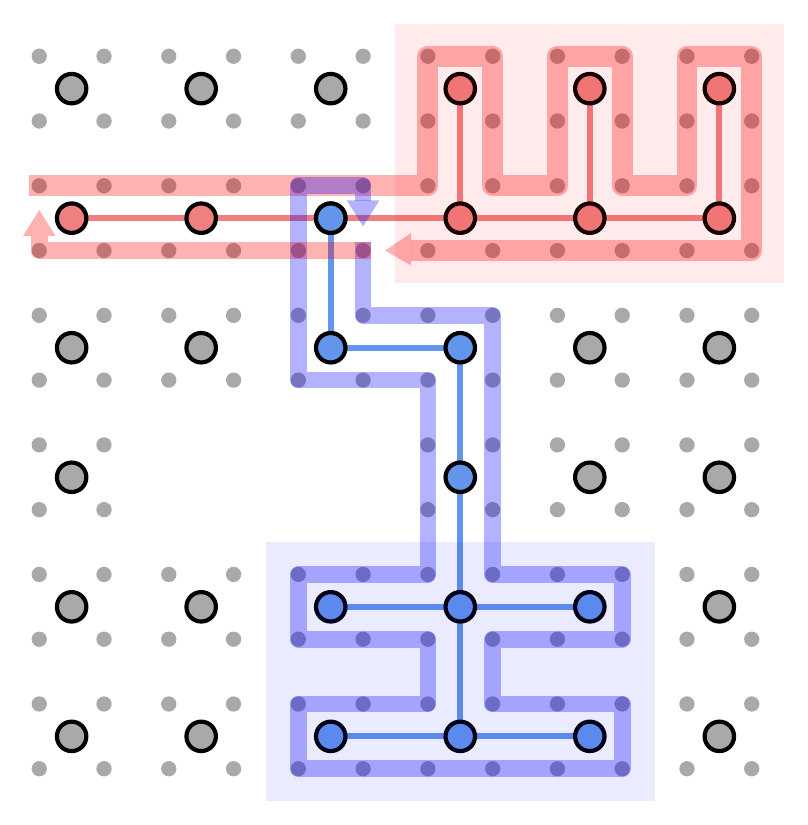}
  \label{fig:method_a_b}
}
\caption{\textbf{Overview of the single-robot path planning strategy.} Red and blue shaded areas indicate the first and second assigned zones $(Z_1,Z_2)$, respectively, along with their corresponding spanning trees $(T_1,T_2)$.
}
\label{fig:method_a}
\end{figure}

We describe the path planning process for a single robot covering multiple zones in a predefined order, as shown in \Cref{fig:method_a}. Starting from its initial position, the robot visits each zone sequentially.
To cover a zone, it first connects from its current position to the closest point in the target zone using a shortest path, then builds a spanning tree covering the entire zone. This path and the spanning tree are combined into a single tree for each zone.
After constructing these trees, the robot generates a complete path by traversing them in order. When a zone is fully covered, it directly moves to the next one, ensuring ordered and continuous coverage across zones.

\begin{algorithm}[t]
\DontPrintSemicolon
\caption{Zone-Wise Tree Construction}
\label{alg:build_trees}

\KwIn{Starting node $u^{\text{start}}$, \quad\quad\quad\quad\quad\quad\quad\quad\quad\quad\quad\quad\quad\quad
List of assigned zones $\mathbf{Z} = [Z_j]_{j=1}^m$}
\KwOut{List of spanning trees $\mathbf{T} = [T_j]_{j=1}^m$}

$\mathbf{T} \gets [\,]$\label{alg:build_trees:1}\;
$u^{\text{anchor}} \gets u^{\text{start}}$\label{alg:build_trees:2}\;
\ForEach{$Z_j \in \mathbf{Z}$}{
    $u^{\text{entry}} \gets \argmin_{u \in Z_j} \textsc{Dist}(u, u^{\text{anchor}})$\label{alg:build_trees:4}\;
    $\widetilde{T} \gets$ shortest path from $u^{\text{anchor}}$ to $u^{\text{entry}}$\label{alg:build_trees:5}\;
    $\widehat{T} \gets$ find MST of $Z_j$\label{alg:build_trees:6}\;
    $\mathbf{T} \gets \mathbf{T} \mathbin\Vert [\widetilde{T} \cup \widehat{T}]$\label{alg:build_trees:7}\;
    $u^{\text{anchor}} \gets \argmin_{u \in \widetilde{T}} \textsc{Dist}(u, u^{\text{entry}})$\label{alg:build_trees:8}\;
}
\Return{$\mathbf{T}$} \label{alg:build_trees:9}
\end{algorithm}

\begin{algorithm}[t]
\DontPrintSemicolon
\caption{Sequential Tree Traversal (\textsc{STT})}
\label{alg:stt}
\KwIn{List of assigned zones $\mathbf{Z} = [Z_j]_{j=1}^m$, \quad\quad\quad\quad\quad\quad
List of spanning trees $\mathbf{T} = [T_j]_{j=1}^m$}
\KwOut{Full coverage path $\pi$}

$\hat{\pi} \gets \text{Tree-Traversal}(\mathbf{T}[0])$\label{alg:stt:1}\;

\uIf{$|\mathbf{Z}|=1$}{ \label{alg:stt:2}
    \Return{$\hat{\pi}$}\label{alg:stt:3}
}
\Else{
    $\pi \gets [\,]$\label{alg:stt:5}\;
    $C \gets \emptyset$\label{alg:stt:6}\;
    \ForEach{$v \in \hat{\pi}$}{
        \uIf{$C = \mathbf{Z}[0]$}{\label{alg:stt:8}
            $\tilde{\pi} \gets \textsc{STT}(\mathbf{Z}[1:], \mathbf{T}[1:])$\label{alg:stt:9}\;
            $\pi\gets\pi\mathbin\Vert\tilde{\pi}$\label{alg:stt:10}\;
        }
        \Else{
            $\pi\gets\pi\mathbin\Vert[v]$\label{alg:stt:12}\;
        }
        \If{$v \in \mathbf{Z}[0]$}{\label{alg:stt:13}
            $C \gets C \cup \{v\}$\label{alg:stt:14}\;
        }
    }
}
\Return{$\pi$}\label{alg:stt:15}
\end{algorithm}

\textbf{Zone-Wise Tree Construction:} \Cref{alg:build_trees} outlines the procedure for constructing a sequence of spanning trees that enables a robot to cover a set of assigned zones $\mathbf{Z} = [Z_j]_{j=1}^m$ in order, starting from its initial position $r$. [Lines \ref{alg:build_trees:1}-\ref{alg:build_trees:2}] The algorithm begins by initializing an empty list to store the resulting trees and setting the anchor node to the starting position. [Line~\ref{alg:build_trees:4}] For each zone in the sequence, the robot identifies the closest node in the zone $Z_j$ to the current anchor $v^{\text{anchor}}$ by computing the shortest distances from the anchor to all nodes using Dijkstra’s algorithm \cite{dijkstra2022note}, and selects this node as the entry point $v^{\text{entry}}$. [Line~\ref{alg:build_trees:5}] The resulting shortest path from the anchor to the entry point is then stored as $\tilde{T}_i$. [Line \ref{alg:build_trees:6}] Then, an internal coverage tree $\hat{T}_i$ is constructed over $Z_j$ using Kruskal’s algorithm \cite{kruskal1956shortest}, yielding a minimum-cost spanning tree. [Line \ref{alg:build_trees:7}] $\tilde{T}_i$ and $\hat{T}_i$ are merged into a single tree, which is then appended to the list $\mathbf{T}$. [Line \ref{alg:build_trees:8}] The anchor node is updated to the node within the connecting path $\tilde{T}_j$ that is nearest to the entry point $v^{\text{entry}}$, allowing the robot to efficiently initiate traversal toward the next zone. [Line \ref{alg:build_trees:9}] The final output is an ordered list of spanning trees, each corresponding to one of the assigned zones which is illustrated in \Cref{fig:method_a_a}.

\textbf{Sequential Tree Traversal:} Once a list of zone-wise spanning trees $\mathbf{T}$ is constructed, the robot generates a complete coverage path by traversing these trees, as outlined in \Cref{alg:stt}. [Line~\ref{alg:stt:1}] The traversal begins with the first tree using a depth-first strategy, following the standard procedure used in CPP problems based on STC. [Lines~\ref{alg:stt:2}–\ref{alg:stt:3}] If there is only one zone, the traversal path $\hat{\pi}$ is directly returned. [Lines~\ref{alg:stt:5}–\ref{alg:stt:6}] Otherwise, the algorithm initializes a coverage buffer $C$ and iteratively appends visited nodes to the final path $\pi$. 
[Lines~\ref{alg:stt:8}–\ref{alg:stt:10}] Once all nodes in the current zone are visited, the algorithm recursively calls itself with the remaining zones and their trees. The returned subpath $\tilde{\pi}$ already contains the connection to the next zone within the pre-constructed trees, and by skipping the current node $v$, the algorithm avoids multiple visits and ensures a continuous transition.
[Line \ref{alg:stt:12}] Otherwise, the node is appended to the path as usual. [Lines~\ref{alg:stt:13}–\ref{alg:stt:14}] Throughout the process, the buffer $C$ is continuously updated to monitor coverage progress within the current zone. [Line \ref{alg:stt:15}] The final output is a full coverage path $\pi$ for the spanning trees $\mathbf{T}$, which is illustrated in \Cref{fig:method_a_b}.

\subsection{Prioritized Zone Assignment for Multiple Robots}\label{subsec:method_2}

The goal of this step is to assign zones, together with their visiting order, to multiple robots in a way that minimizes the priority-weighted latency objective defined in \Cref{eq:2}. We formulate this as a variant of the multi-depot $k$-TRP, where each robot’s initial position acts as a depot and each zone serves as a target to be visited \cite{bruni2022multi}. The objective is to minimize the total priority-weighted arrival time of all zones across robots. However, obtaining an exact optimal solution through integer programming (IP) is computationally intractable for large-scale scenarios, as noted in prior work \cite{bruni2022multi}. To address this, we adopt a practical approach combining a greedy assignment with a local search refinement.

\begin{algorithm}[t]
\DontPrintSemicolon
\caption{Greedy Zone Assignment}
\label{alg:gza}
\KwIn{Robot set $I$ with initial positions $\{r_i\}_{i\in I}$,\quad\quad\quad\quad Zones $Z=\{Z_j\}_{j=1}^n$, Priority weights $\{w_j\}_{j=1}^n$}
\KwOut{Ordered zone assignments $\{\mathbf{Z}_i\}_{i\in I}$}

$\mathcal{U} \gets Z$ \tcp{unassigned zones} \label{alg:gza:1}
\ForEach{$i \in I$}{
    $\mathbf{Z}_i \gets ()$; $T_i \gets 0$\label{alg:gza:3}
}

\tcp{Compute traversal costs}
\ForEach{$Z_j \in Z$}{\label{alg:gza:4}
    $c(Z_j) \gets \text{MST cost of } Z_j$\;\label{alg:gza:5}
}

\ForEach{$i \in I$, $Z_j \in Z$}{\label{alg:gza:6}
    $c(i,Z_j) \gets \min_{v \in Z_j}\textsc{Dist}(r_i,v)$\;
}
\ForEach{$Z_j,Z_{j'} \in Z$ with $j < j'$}{
    $c(Z_j,Z_{j'}) \gets \min_{v \in Z_j,v' \in Z_{j'}}\textsc{Dist}(v,v')$\;
    $c(Z_{j'},Z_j) \gets c(Z_j,Z_{j'})$\label{alg:gza:10}
}

\tcp{Greedy assignment loop}
\While{$\mathcal{U} \neq \emptyset$}{\label{alg:gza:11}
    $\Delta_{\min} \gets +\infty$\;
    \ForEach{$Z_j \in \mathcal{U}$}{
        \ForEach{$i \in I$}{
            \uIf{$\mathbf{Z}_i = \emptyset$}{
                $\Delta \gets w_j \cdot (T_i + c(i,Z_j) + c(Z_j))$\;
            }
            \Else{
                $Z_{\text{end}} \gets$ last zone in $\mathbf{Z}_i$\;
                $\Delta \gets w_j \cdot (T_i + c(Z_{\text{end}}, Z_j) + c(Z_j))$\;
            }
            \If{$\Delta < \Delta_{\min}$}{
                $\Delta_{\min} \gets \Delta$; $(i^*, Z_{j^*}) \gets (i, Z_j)$\;
            }
        }
    }
    $\mathbf{Z}_{i^*} \gets \mathbf{Z}_{i^*} \Vert [Z_{j^*}]$\;
    \uIf{$|\mathbf{Z}_{i^*}| = 1$}{
        $T_{i^*} \gets T_{i^*} + c(i^*, Z_{j^*}) + c(Z_{j^*})$\;
    }
    \Else{
        $Z_{\text{prev}} \gets$ previous last zone in $\mathbf{Z}_{i^*}$\;
        $T_{i^*} \gets T_{i^*} + c(Z_{\text{prev}}, Z_{j^*}) + c(Z_{j^*})$\;
    }
    $\mathcal{U} \gets \mathcal{U}\setminus \{Z_{j^*}\}$\;\label{alg:gza:28}
}
\Return{$\{\mathbf{Z}_i\}_{i\in I}$}\label{alg:gza:29}
\end{algorithm}

\textbf{Greedy Zone Assignment:} \Cref{alg:gza} outlines the procedure for assigning ordered zones to multiple robots, minimizing the total priority-weighted latency. [Lines~\ref{alg:gza:1}-\ref{alg:gza:3}] The algorithm begins by initializing an empty sequence $\mathbf{Z}_i$ and traversal time $T_i$ for each robot, with all zones marked as unassigned. [Lines~\ref{alg:gza:4}-\ref{alg:gza:5}] For each zone, the internal cost of traveling within the zone, $c(Z_j)$, is computed using Kruskal’s algorithm. [Lines~\ref{alg:gza:6}-\ref{alg:gza:10}] Depot-to-zone traversal costs $c(i, Z_j)$ are computed via Dijkstra’s algorithm for each robot-zone pair, while inter-zone traversal costs $c(Z_j, Z_{j'})$ are optimistically set as the minimum distance between any node pairs from the two zones. [Lines~\ref{alg:gza:11}-\ref{alg:gza:28}] Then, a greedy assignment loop iteratively selects the robot-zone pair $(i, Z_j)$ that minimizes the incremental priority-weighted latency $\Delta$, considering both travel and internal zone costs. If a robot has no prior assignment, the cost from its depot is used; otherwise, the cost from the last assigned zone is considered. The chosen zone $Z_{j^*}$ is appended to robot $i^*$'s sequence $\mathbf{Z}_i$, and its traversal time $T_{i^*}$ is updated. The selected zone is removed from the unassigned set $\mathcal{U}$. This process continues until all zones are assigned. [Line ~\ref{alg:gza:29}] The final output is the ordered zone assignments ${\mathbf{Z}_i}$ for all robots, which serve as input for subsequent single-robot path planning described in \Cref{subsec:method_1}.

\textbf{Local Search Refinement:} After obtaining an initial assignment of ordered zone sequences $\{\mathbf{Z}_i\}_{i \in I}$, we refine assignments using local search to further reduce the total priority-weighted latency. At each iteration, we define a local neighborhood by applying one of two types of perturbations:
\begin{enumerate}
    \item \textbf{Move operator}: Randomly select a zone $Z_j$ from a robot $i$'s sequence and move it to a random position within a random robot $i'$'s sequence.
    \item \textbf{Swap operator}: Randomly select two zones $Z_j$ and $Z_{j'}$ assigned to possibly different robots, and directly swap their assignments and respective positions.
\end{enumerate}
After each modification, we recompute the total objective for the new assignment. If the new configuration results in a reduced cost, it is accepted; otherwise, it is discarded. This process iteratively explores the neighborhood around the current solution, gradually improving it.

\subsection{Residual Area Coverage with a Steiner Tree}\label{subsec:method_3}

\begin{figure}[t!]
\centering
\includegraphics[width=0.8\columnwidth]{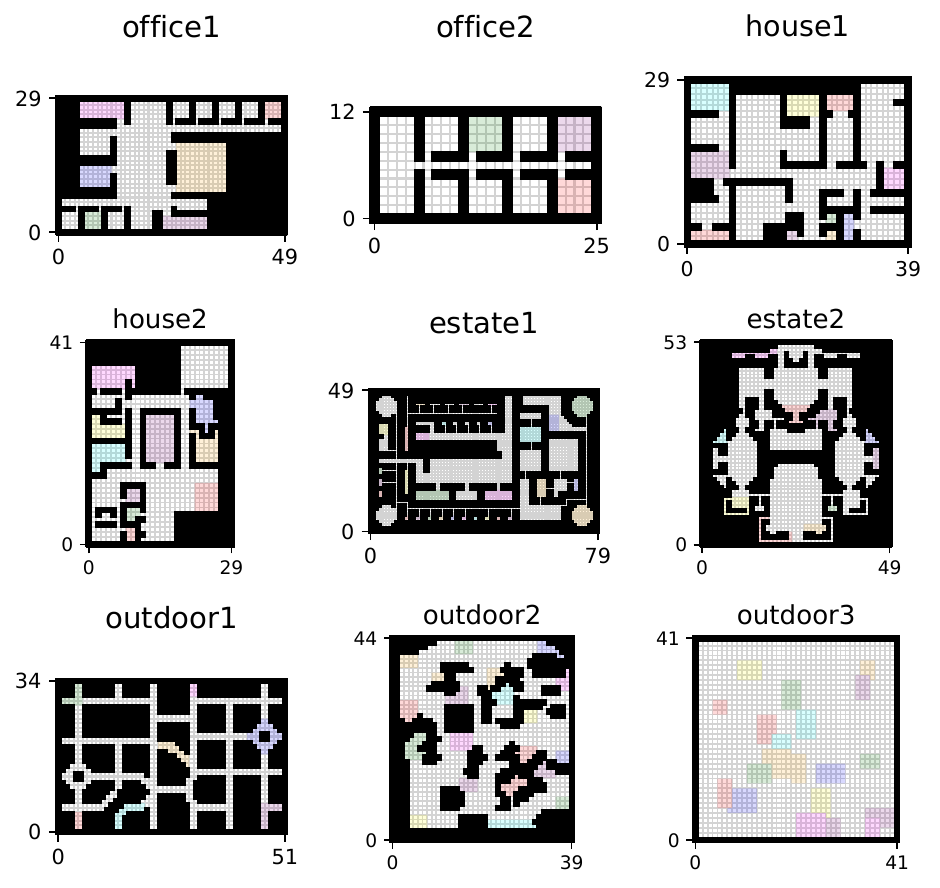}
\vspace{-10pt}
\caption{\textbf{PA-MCPP map instances.} Black, white, and colored squares represent obstacles, normal terrain, and prioritized zones, respectively. Different colors indicate different zones.}
\label{fig:exp_map}
\end{figure}

After each robot completes its prioritized zone coverage by traversing the constructed trees $\{\mathbf{T}_i\}_{i\in I}$ in the hypergraph $H = (V_h, E_h)$, we identify the set of already covered hypervertices. This set is denoted as $\mathcal{V}_{\text{covered}} = \bigcup_{i\in I} V_h(\mathbf{T}_i)$, where $V_h(\mathbf{T}_i)$ represents the hypervertices included in robot $i$’s coverage tree.
The remaining uncovered region is $\mathcal{V}_{\text{res}} = V_h \setminus \mathcal{V}_{\text{covered}}$.
To efficiently cover $\mathcal{V}_{\text{res}}$, we construct a Steiner tree $T_S$ on the hypergraph $H$, where the terminal set is the set of residual hypervertices $\mathcal{V}_{\text{res}}$. A Steiner tree is the MST over the terminal set, allowing the inclusion of additional non-terminal vertices if this reduces the total connection cost.

We then generate a single traversal path on $T_S$ and partition it among robots using an MSTC$^*$ \cite{tang2021mstc}. The partitioning aims to minimize $\max_{i \in I}\big(C_i^{(1)} + C_i^{(2)}\big)$, where $C_i^{(1)}$ is the cost from phase one and $C_i^{(2)}$ is the residual cost after partitioning. This min–max allocation ensures a balanced workload and minimal makespan by directly accounting for each robot’s prior effort. The resulting residual path is concatenated after the prioritized zone path to form the final path for each robot.

\section{Experiments}
    \label{sec:experiments}

\begin{table*}[t]
\centering
\caption{Performance comparison with baselines on different map instances}
\vspace{-5pt}
\resizebox{0.87\textwidth}{!}{
\begin{tabular}{c|c|c|c|c|ccc|ccc}
\toprule
\multirow{2}{*}{\textbf{Instance}} &
\multirow{2}{*}{\textbf{Map Size}} &
\textbf{\# of} &
\textbf{\# of} &
\multirow{2}{*}{\textbf{Weighted}} &
\multicolumn{3}{c|}{\textbf{Zone Coverage Latency ($\downarrow$)}} &
\multicolumn{3}{c}{\textbf{Makespan ($\downarrow$)}}\\
&
&\textbf{Zone}
&\textbf{Robot}
&& MFC & MSTC$^*$ & PA-MCPP & MFC & MSTC$^*$ & PA-MCPP \\
\midrule

\texttt{office1} & 49$\times$29 & 6 & 10 & \xmark &431.1 ± 26.0 & 443.9 ± 50.5 & 283.7 ± 5.5 & 121.7 ± 10.2 & 96.7 ± 3.9 & 127.9 ± 5.2 \\

\texttt{office2} & 25$\times$12 & 3 & 3 & \xmark &104.2 ± 13.1 & 109.5 ± 17.5 & 65.7 ± 4.8 & 64.8 ± 8.8 & 47.7 ± 1.9 & 54.1 ± 3.2 \\

\texttt{house1} & 39$\times$29 & 10 & 5 & \xmark & 1182.3 ± 152.9 & 1138.3 ± 226.8 & 339.9 ± 15.0 & 226.5 ± 19.8 & 183.1 ± 3.6 & 193.2 ± 3.9 \\

\texttt{house2} & 29$\times$41 & 10 & 10 & \xmark &1065.8 ± 145.8 & 1066.2 ± 85.4 & 455.5 ± 17.1 & 213.6 ± 13.6 & 157.8 ± 3.3 & 168.5 ± 5.4 \\

\texttt{estate1} & 79$\times$49 & 30 & 5 & \xmark & 7623.7 ± 834.9 & 7480.0 ± 772.8 & 1964.3 ± 85.6 & 566.5 ± 57.1 & 433.1 ± 14.1 & 508.4 ± 13.5 \\

\texttt{estate2} & 49$\times$53 & 10 & 10 & \xmark & 1026.4 ± 157.0 & 995.1 ± 102.3 & 248.6 ± 17.6 & 200.6 ± 27.1 & 150.8 ± 5.5 & 153.7 ± 6.9 \\

\texttt{outdoor1} & 51$\times$34 & 7 & 10 & \cmark &443.9 ± 52.8 & 446.7 ± 46.0 & 158.9 ± 6.3 & 125.3 ± 9.7 & 88.9 ± 3.5 & 89.6 ± 3.7 \\

\texttt{outdoor2} & 39$\times$44 & 20 & 15 & \cmark &1793.9 ± 75.0 & 1698.7 ± 111.6 & 558.2 ± 22.6 & 171.5 ± 10.8 & 128.2 ± 3.5 & 134.4 ± 4.6 \\

\texttt{outdoor3} & 41$\times$41 & 20 & 20 & \cmark &7982.4 ± 1002.8 & 7690.2 ± 391.4 & 1616.2 ± 35.3 & 661.3 ± 79.7 & 545.1 ± 4.4 & 570.1 ± 5.5 \\

\midrule
\midrule

\multicolumn{5}{r|}{\textbf{Avg. Improvement vs.~MSTC$^*$ ($\%$)}}
& -1.1 ± 3.3 & - & 62.5 ± 14.4 & -31.2 ± 6.0 & - & -9.7 ± 9.4 \\

\bottomrule
\end{tabular}
}
\vspace{-0.3cm}
\label{table:exp_comb}
\end{table*}

\begin{figure}[t!]
\centering
\subfloat{
  \includegraphics[width=0.20\textwidth]{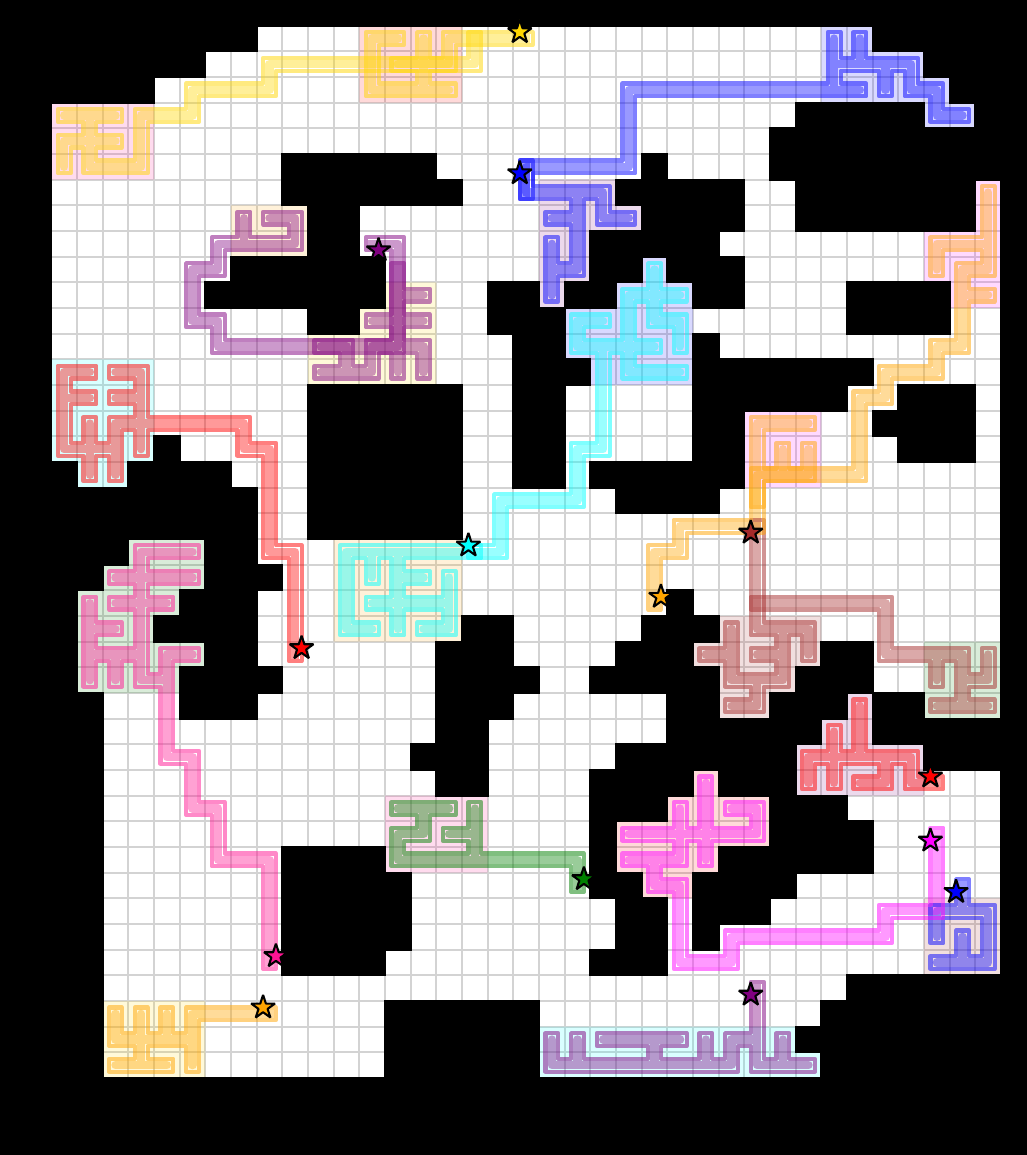}
  \label{fig:exp1_a}
}
\subfloat{
  \includegraphics[width=0.20\textwidth]{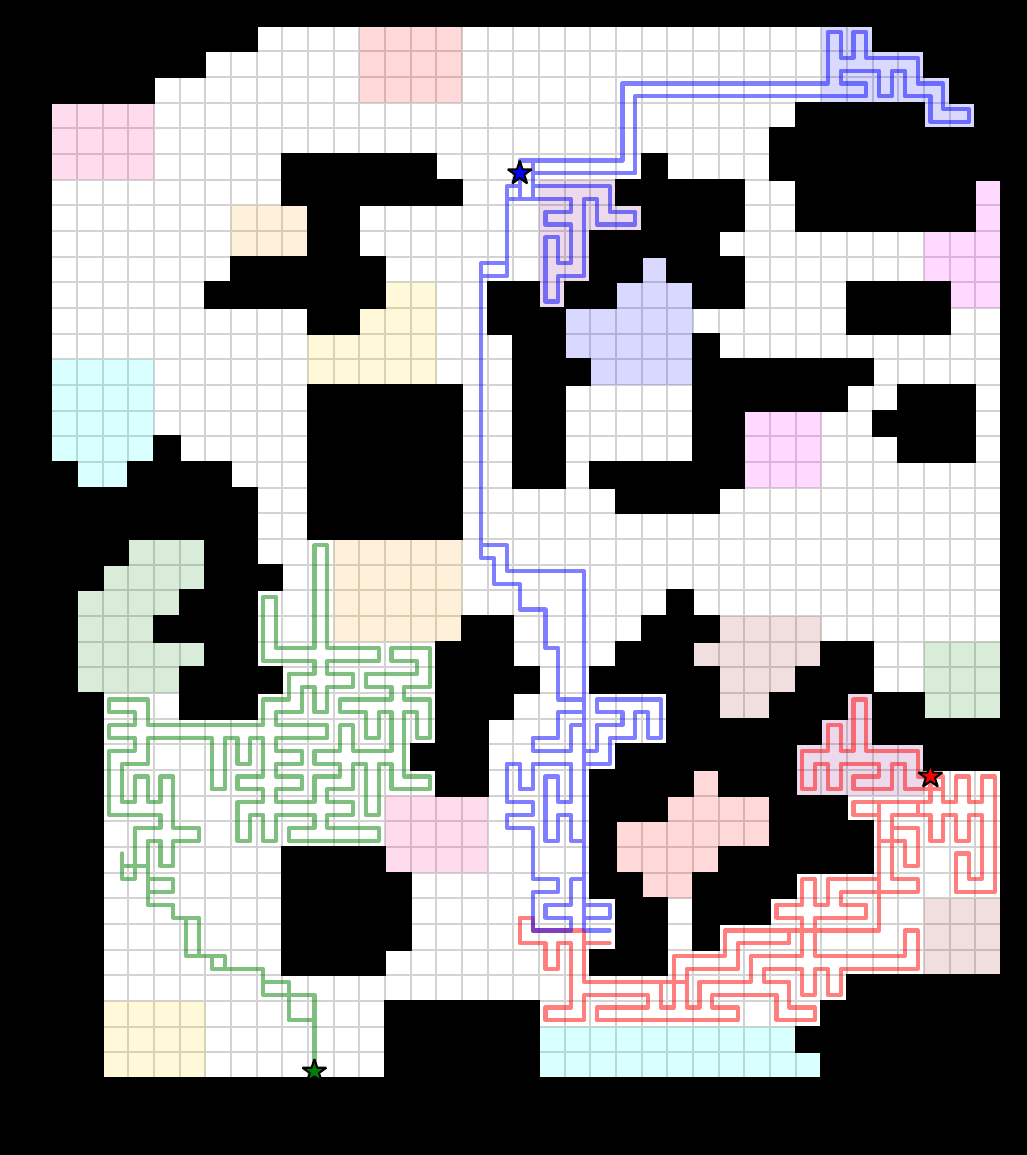}
  \label{fig:exp1_b}
}
\caption{\textbf{Visualization of zone assignments and robot coverage paths.} Left: Matched priority zones with corresponding spanning trees and assigned paths. Right: Full coverage paths for Robot 1 (red), Robot 3 (green), and Robot 15 (blue), showing prioritized zone coverage followed by residual area coverage.}
\label{fig:exp1}
\end{figure}

In the following experiments, we aim to validate four key hypotheses. (1) Can our proposed method effectively solve the PA-MCPP problem? (2) Does our method scale well with the number of robots? (3) Can we control the behavior of the proposed method by adjusting the priority weights assigned to zones? (4) Can our greedy zone assignment combined with local search produce sufficiently high-quality assignments? Note that in all experiments, we conducted 10 trials using random initial robot positions on non-overlapping normal terrain to ensure robustness and fairness of the results, following the experimental protocol in the previous work \cite{tang2025large}. All experiments were performed on a Linux workstation equipped with an AMD Ryzen Threadripper 3970X 32-core processor.

\subsection{Performance Comparison}

This subsection evaluates whether the proposed PA-MCPP lowers priority-weighted zone coverage latency while keeping the makespan competitive. For comparison, we consider several established multi-robot coverage algorithms as baselines: MFC \cite{zheng2005multi} and MSTC$^*$ \cite{tang2021mstc}. 
Although these methods are not explicitly designed for priority objectives, they serve as useful makespan references. We computed the zone coverage latency by post-processing their generated paths.
Including these baselines allows us to verify that our method does not significantly compromise the makespan while optimizing the zone coverage latency. All baseline implementations are based on publicly available code to ensure fairness and reproducibility\footnote{\url{https://github.com/reso1/MSTC_Star}}.

We evaluate the proposed method on nine different map layouts designed to represent a wide range of scenarios, including \texttt{estate}, \texttt{house}, \texttt{office}, and \texttt{outdoor} environments, as illustrated in \Cref{fig:exp_map}. Each map is configured with varying numbers of robots and zones, as noted in \Cref{table:exp_comb} All zones are assigned uniform priority weights of 1. In the third row of map instances, vertex weights are randomly sampled from a uniform distribution $\mathcal{U}(0.8, 1.2)$ to introduce variability and better reflect realistic, non-uniform environments. 
\Cref{table:exp_comb} shows the performance comparison with baselines on different map instances. 
Across all instances, PA-MCPP reduces latency by an average of $62.5\pm14.4\%$ relative to MSTC$^*$, while maintaining a comparable makespan with only $9.7\pm9.4\%$ overhead. Overall, PA-MCPP achieves the desired trade-off, substantially reducing the delay time to complete high-priority zones while introducing only a modest increase in makespan.

\Cref{fig:exp1} illustrates the priority-aware coverage assignments and the final paths of certain robots. In \Cref{fig:exp1_a}, each priority zone is properly assigned to the robot to minimize travel delay, after which coverage paths are generated by traversing the corresponding spanning trees sequentially.
\Cref{fig:exp1_b} depicts the complete coverage trajectories for three representative robots: Robot 1 (red), Robot 3 (green), and Robot 15 (blue). Robot 3 is not assigned any priority zone and thus immediately proceeds to its residual coverage region. Robot 1 has a single priority zone located directly adjacent to its starting position, which it covers first before continuing to residual areas. Robot 15 is assigned two priority zones, which it completes sequentially before transitioning to its remaining coverage region. These results indicate that our zone assignment strategy not only considers proximity but also accounts for each robot’s initial position and the size of the priority zones, thereby reducing travel overhead and enabling early completion of high-priority areas without significantly impacting overall coverage efficiency.

\subsection{Sensitivity Analysis}\label{subsec:sensitivity}

In this subsection, we validate whether our method can scale effectively with the number of robots and whether its behavior can be controlled by adjusting the priority weights of zones. All experiments are conducted on the \texttt{estate1} with $5$ robots as the default configuration, unless otherwise specified.

\begin{figure}[t!]
\centering
\includegraphics[width=0.38\textwidth]{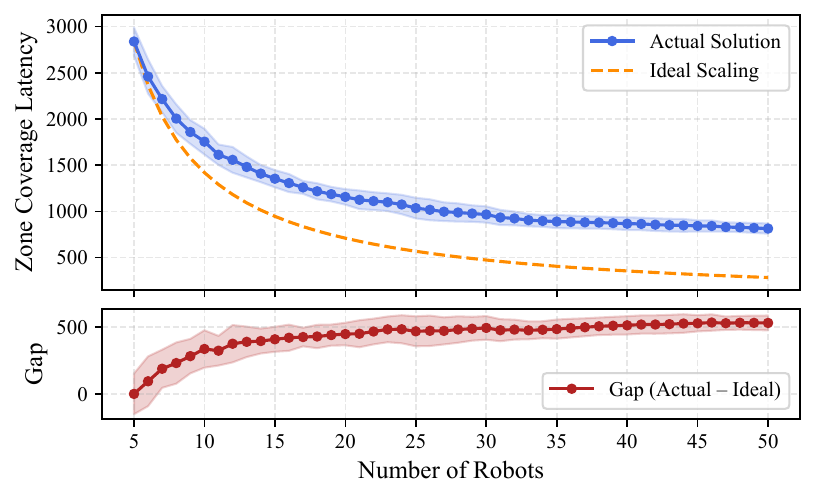}
\vspace{-8pt}
\caption{\textbf{Sensitivity analysis on the number of robots.} Top: zone coverage latency versus the number of robots, along with the ideal scaling baseline. Bottom: the gap between the actual solution and the ideal baseline.}
\vspace{-20pt}
\label{fig:exp3_robot}
\end{figure}

\Cref{fig:exp3_robot} shows the relationship between zone coverage latency and the number of robots. We observe that the overall latency decreases as the number of robots increases, indicating that our algorithm can effectively utilize additional robots to improve coverage speed. However, when comparing the performance to an ideal scaling baseline (calculated as $y_{\text{sol}}^5 \times \frac{5}{\text{num\_robot}}$, where $y_{\text{sol}}^5$ denotes the zone coverage latency of actual solution with five robots), we notice that the performance gap increases with more robots. This ideal scaling assumes perfect parallelization and no additional coordination overhead, which is rarely achievable in practice. The widening gap suggests that while our method remains effective at larger scales, its efficiency diminishes slightly due to increased complexity in robot coordination and zone assignments.

\begin{figure}[t!]
\centering
\subfloat[Zone $1$ (the smallest zone)]{
  \includegraphics[width=0.20\textwidth]{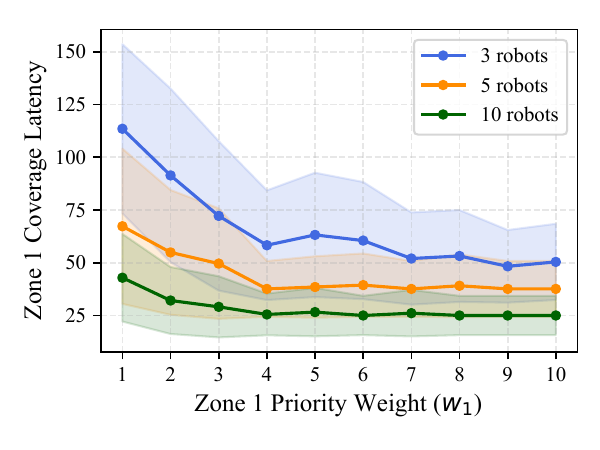}
  \label{fig:exp3_sen_zone_a}
}
\subfloat[Zone $27$ (the largest zone)]{
  \includegraphics[width=0.20\textwidth]{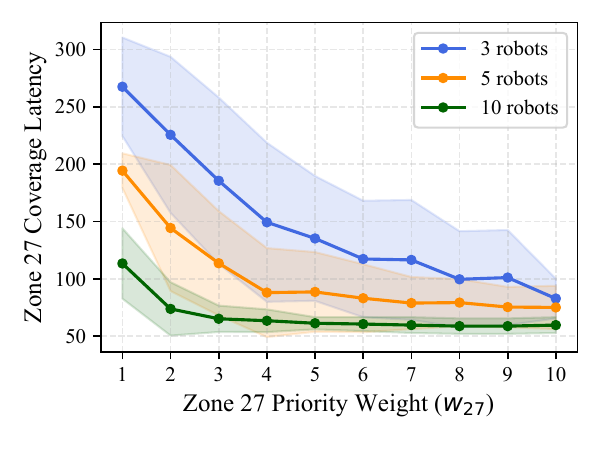}
  \label{fig:exp3_sen_zone_b}
}
\caption{\textbf{Sensitivity analysis on priority weights.} Zone coverage latency under different priority weights and numbers of robots.}
\label{fig:exp3_sen_zone}
\end{figure}

\Cref{fig:exp3_sen_zone} evaluates whether zone coverage can be expedited by increasing priority weights. We perform a sensitivity analysis on the smallest zone (Zone 1) and the largest zone (Zone 27) in the map, using different numbers of robots. For Zone 1, we observe that coverage time decreases rapidly as its priority weight increases, but it converges quickly, showing diminishing returns beyond a certain weight. In contrast, for Zone 27, which is significantly larger, the convergence is more gradual, and the benefit of increasing the priority weight depends more strongly on the number of robots. Also, these results suggest that our method allows flexible control over zone coverage behavior by adjusting priority weights, validating that the algorithm can adapt to user-specified preferences.

\subsection{Evaluation of Zone Assignment Strategies}

\begin{figure}[t!]
\centering
\subfloat{
  \includegraphics[width=0.22\textwidth]{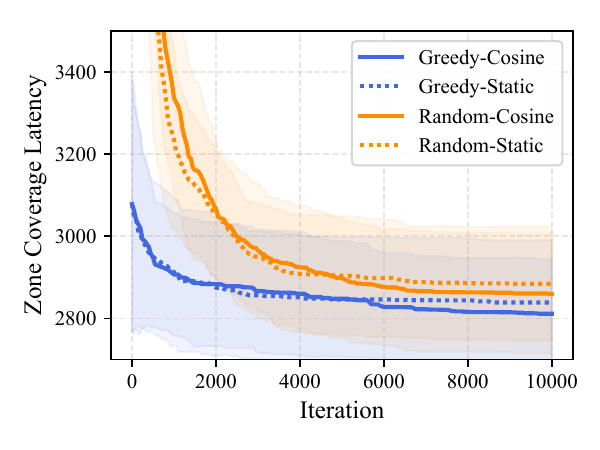}
  \label{fig:exp4_ls_a}
}
\subfloat{
  \includegraphics[width=0.22\textwidth]{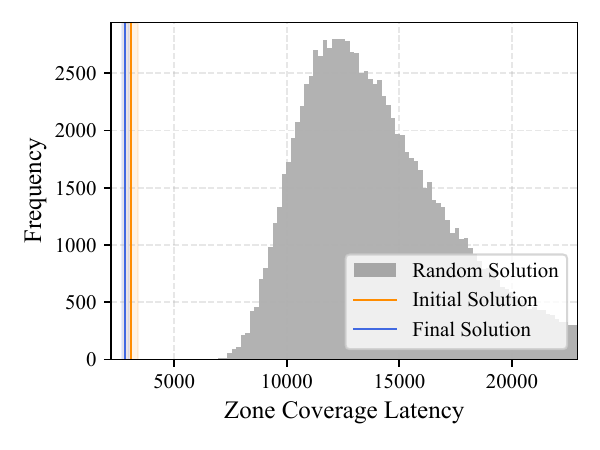}
  \label{fig:exp4_ls_b}
}
\vspace{-10pt}
\caption{\textbf{Results of zone assignment optimization.} Left: Zone coverage latency over local search iterations for different initialization and operator selection strategies. Right: Distribution of zone coverage latency for random assignments, with vertical lines indicating the initial greedy solution and the final solution after local search.}
\label{fig:exp4_ls}
\vspace{-10pt}
\end{figure}

In this subsection, we validate whether the proposed greedy zone assignment combined with local search can produce high-quality assignments for multiple robots. Experiments are conducted on the \texttt{estate1} layout, with five robots as in \Cref{subsec:sensitivity}. The left plot in \Cref{fig:exp4_ls} shows the change in zone coverage latency over local search iterations for different initialization and operator selection strategies. We compare greedy against random initialization and evaluate two operator variants: cosine and static schedules. In the cosine schedule, we set the period to $10\%$ of the total number of iterations, resulting in each operator’s selection probability oscillating smoothly between $1.0$ and $0.0$ throughout each period. In the static schedule, the move or swap operator is selected uniformly at random. Greedy initialization consistently provides a better starting point than random initialization. As the number of local search iterations increases, the solutions further improve and eventually converge. However, greedy initialization consistently produces better final solutions than random initialization. Also, the cosine schedule achieves better results than the static schedule for both initialization methods.

\begin{figure}[t!]
\centering
\includegraphics[width=0.38\textwidth]{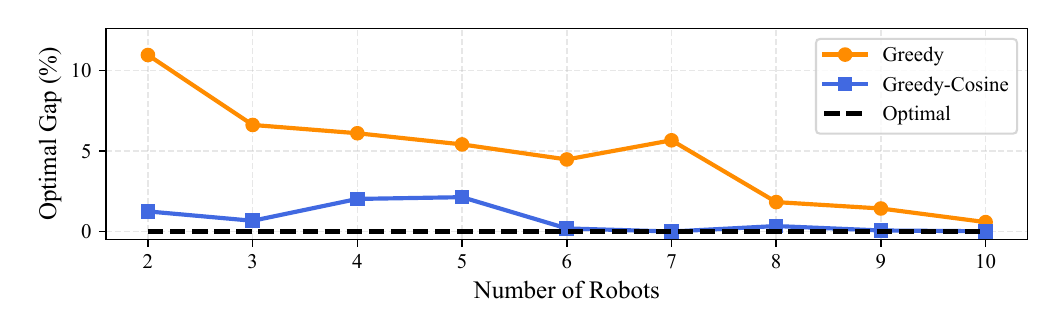}
\vspace{-12pt}
\caption{\textbf{Comparison with the optimal IP solution on zone assignment.} Optimality gap of zone coverage latency versus the number of robots for initial greedy solution, the final solution after local search, and the optimal IP baseline on \texttt{outdoor1}.}
\label{fig:exp_optimal_ip2}
\end{figure}

\begin{table}[t]
\centering
\caption{Comparison of Max-to-Mean Ratio for workload balancing}
\resizebox{0.7\columnwidth}{!}{%
\begin{tabular}{c|cc|c}
\toprule
\multirow{2}{*}{\textbf{\# of Robots}} &
\multicolumn{3}{c}{\textbf{Max-to-Mean Ratio ($\downarrow$)}} \\
& MSTC$^*$ & PAMCPP & Diff. ($\Delta$) \\
\midrule
5  & 1.01 $\pm$ 0.01 & 1.01 $\pm$ 0.0 & -0.01 $\pm$ 0.01 \\
10 & 1.04 $\pm$ 0.04 & 1.05 $\pm$ 0.03 &  0.01 $\pm$ 0.05\\
15 & 1.19 $\pm$ 0.13 & 1.20 $\pm$ 0.08 &  0.01 $\pm$ 0.15\\
20 & 1.21 $\pm$ 0.15 & 1.27 $\pm$ 0.18 &  0.06 $\pm$ 0.23\\
\bottomrule
\end{tabular}
}
\vspace{-10pt}
\label{table:maxmean_ratio}
\end{table}

The right plot in \Cref{fig:exp4_ls} shows the distribution of zone coverage latency for random assignments, illustrating the substantial performance gap between random and our solutions. The vertical lines indicate the latency of the initial greedy assignment and the final solution after local search refinement. These results demonstrate the effectiveness of combining a strong initial assignment with local search, supporting the ability of our method to find high-quality zone assignments.

\Cref{fig:exp_optimal_ip2} compares the proposed solutions with the optimal IP baseline \cite{bruni2022multi}. The results show that combining greedy initialization with local search achieves near-optimal performance, demonstrating the effectiveness of the proposed algorithm. As the number of robots increases, the optimality gap narrows because each robot is responsible for fewer zones. In such cases, a simple greedy assignment becomes nearly optimal, as most robots are assigned to the closest zone. Due to computational complexity in solving the IP formulation, this comparison was conducted only on maps (e.g., \texttt{outdoor1}) with a relatively small number of zones.

\subsection{Evaluation of Robot Workload Balancing}

To quantitatively assess workload balancing, we compute the Max-to-Mean Ratio (MMR) of the path lengths assigned to individual robots, defined as the ratio between the maximum individual workload and the mean workload across all robots, where values closer to 1 indicate better balance. \Cref{table:maxmean_ratio} reports the mean and standard deviation of MMR for different team sizes.
Across all cases, the MMR values of PA-MCPP remain comparable to those of MSTC$^*$, with differences ($\Delta$) within statistical variation. For larger teams (e.g., 20 robots), the variance increases ($0.06 \pm 0.23$), indicating that some trials exhibit less balanced workloads. This variability arises from the inherent characteristics of the MSTC$^*$, where maintaining perfectly balanced workloads becomes increasingly difficult as the number of robots grows. Our integration introduces only marginal additional imbalance beyond this baseline behavior. Overall, PA-MCPP maintains workload balance comparable to MSTC$^*$ while achieving its primary performance gains.

\subsection{Computational Time Analysis}

\begin{figure}[t!]
\centering
\includegraphics[width=0.38\textwidth]{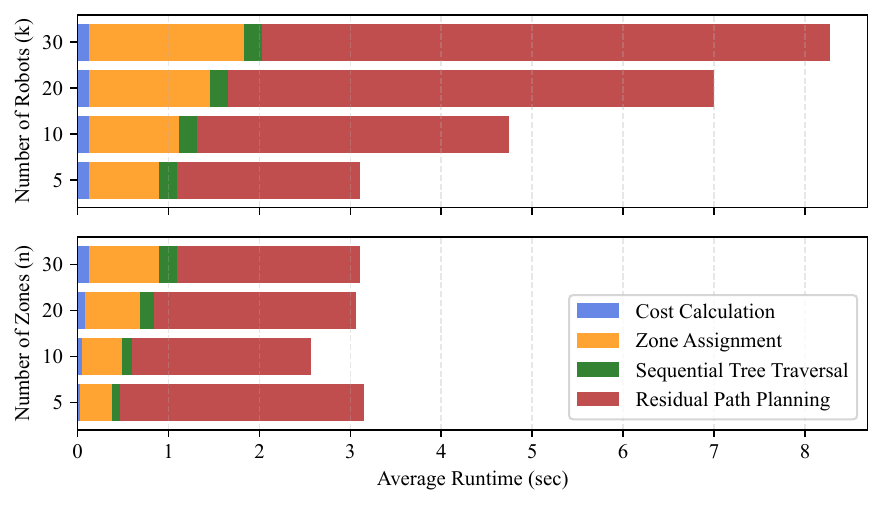}
\vspace{-12pt}
\caption{\textbf{Average runtime breakdown by algorithmic stage.} Top: Mean execution time for varying numbers of robots $k$, with the number of zones fixed at $30$. Bottom: Mean execution time for varying numbers of zones $n$, with the number of robots fixed at $5$.}
\label{fig:exp_runtime}
\end{figure}

\Cref{fig:exp_runtime} presents the average runtime breakdown of the proposed algorithm into four major computational stages. Across all tested configurations, residual path planning constitutes the dominant portion of the computation time, with its contribution becoming more pronounced as the number of robots increases. The runtime of cost calculation, zone assignment, and sequential tree traversal exhibit gradual growth with increasing $k$ or $n$, reflecting the additional pairwise computations and assignment complexity. However, residual path planning does not scale monotonically with $n$. When $n$ is small (e.g., $n = 5$), the number of uncovered nodes after the initial coverage is relatively large, causing the residual path planning step to require substantially more processing than for $n = 10$ or $20$. For larger $n$ (20–30), this effect saturates and results in comparable planning times.
Overall, the number of robots has a greater impact on computation time than the number of zones, mainly due to the larger branching factor and increased complexity in assignment and traversal.

\section{Conclusion and Future Work}
    \label{sec:conclusion}

In this paper, we proposed PA-MCPP, a two-phase framework that considers both prioritized zone importance and overall coverage efficiency. The method combines greedy zone assignment with local search refinement, followed by residual area coverage to ensure completeness. Experiments show that it effectively reduces priority-weighted latency while maintaining competitive makespan, demonstrating scalability and controllability through sensitivity analyses.
However, a primary limitation of the current framework is the locally updated anchor strategy, which may lead to suboptimal zone connections due to its myopic design and the one-to-one assignment between zones and robots, which may hinder scalability in environments that contain large but few prioritized zones. As future work, we plan to explore more flexible assignments that allow multiple robots to share a single zone.

\printbibliography

\end{document}